\documentclass[letterpaper, 10 pt, conference]{ieeeconf}  % Comment this line out if you need a4paper

\IEEEoverridecommandlockouts                              % This command is only needed if 
\usepackage[utf8]{inputenc}
\usepackage[english]{babel}
\usepackage{amsmath}
\usepackage{amssymb}
\usepackage{graphicx}
\usepackage{subcaption}
\usepackage{hyperref}
\usepackage{url}
\usepackage{multirow}
\usepackage[dvipsnames]{xcolor}

\title{\LARGE \bf ORBSLAM-Atlas: a robust and accurate multi-map system}
\author{Richard Elvira, Juan D. Tardós and J.M.M. Montiel
\thanks{This work was supported in part by the Spanish government under grants PGC2018-096367-B-I00 and DPI2017-91104-EXP, by the Aragón government under grant DGA\_T45-17R, and by Huawei under grant HF2017040003.}% <-this % stops a space
\thanks{The authors are with Instituto de Investigación en ingeniería de Aragón (I3A), Universidad de Zaragoza, Spain {\tt\small richard@unizar.es;  tardos@unizar.es; josemari@unizar.es;}}%
}

\begin{document}

\onecolumn
\begin{center}
This paper has been accepted in 2018 IEEE/RSJ International Conference on Intelligent Robots and Systems (IROS)
\newline 

\end{center}
\textcopyright 2019 IEEE. Personal use of this material is permitted. Permission from IEEE must be
obtained for all other uses, in any current or future media, including
reprinting/republishing this material for advertising or promotional purposes, creating new
collective works, for resale or redistribution to servers or lists, or reuse of any copyrighted
component of this work in other works.

\newpage
\twocolumn

\maketitle
\thispagestyle{empty}
\pagestyle{empty}

\begin{abstract}
We propose ORBSLAM-Atlas, a system able to handle an unlimited number of disconnected sub-maps, that includes a robust map merging algorithm able to detect sub-maps with common regions and seamlessly fuse them. The outstanding robustness and accuracy of ORBSLAM are due to its ability to detect wide-baseline matches between keyframes, and to exploit them by means of non-linear optimization, however it only can handle a single map. ORBSLAM-Atlas brings the wide-baseline matching detection and exploitation to the multiple map arena. The result is a SLAM system significantly more general and robust, able to perform multi-session mapping. If tracking is lost during exploration, instead of freezing the map, a new sub-map is launched, and it can be fused with the previous map when common parts are visited. Our criteria to declare the camera lost contrast with previous approaches that simply count the number of tracked points, we propose to discard also inaccurately estimated camera poses due to bad geometrical conditioning. As a result, the map is split into more accurate sub-maps, that are eventually merged in a more accurate global map, thanks to the multi-mapping capabilities.

We provide extensive experimental validation in the EuRoC datasets, where ORBSLAM-Atlas obtains accurate monocular and stereo results in the difficult sequences where ORBSLAM failed. We also build global maps after multiple sessions in the same room, obtaining the best results to date, between 2 and 3 times more accurate than competing multi-map approaches. We also show the robustness and capability of our system to deal with dynamic scenes, quantitatively in the EuRoC datasets and qualitatively in a densely populated corridor where camera occlusions and tracking losses are frequent.

\end{abstract}

%-------------------------------------------

\section{Introduction}
SLAM (Simultaneous Localization and Mapping) algorithms are able to build a map from sensor readings, and simultaneously estimate the sensor localization within the map. Cameras are particularly interesting sensors because of the unique combination of geometry and semantics they provide. In this case, the algorithms are dubbed V-SLAM (Visual SLAM), in this work we focus on the purely visual monocular and stereo sensors. We focus on  keyframe and feature point SLAM methods because of their relocalization and place recognition performance, displayed in their capability to build up to city block size maps robustly.

More specifically we build on top of the reference system ORBSLAM \cite{mur2015orb, mur2017orb,mur2017visual}. If compared with visual odometry methods \cite{qin2018vins, qin2019general, delmerico2018benchmark,Engel-et-al-pami2018,forster2017svo}, ORBSLAM can perform far more accurately especially if the same area is revisited. The ORBSLAM accuracy comes from non-linear bundle adjustment (BA) in which the observations of the same map point come from widely separated keyframes. On the one hand, ORBSLAM is able to robustly detect matches between keyframes even if they are widely separated in time, even in the extreme case of loop closure. ORBSLAM is able to make the most of these abundant high parallax re-observations by an intertwining of elementary mapping stages: ORB matching, DBoW2 place recognition, pose graph optimization, local BA, global BA, and map management. The map management includes creation, deletion, and merging of map points and keyframes. However, it can only handle a single map, which provokes a total failure in exploratory trajectories if tracking is lost, and prevents multi-session mapping. 

We propose the ORBSLAM-Atlas system, a generalization of ORBSLAM to the multiple map case. Our main contributions are:
\begin{itemize}
\item A multi-map representation that we call \emph{atlas}, that handle an unlimited number of sub-maps. 
The atlas has a unique DBoWs database of keyframes for all the sub-maps, which allows efficient multi-map place recognition. 
\item Algorithms for all the multi-mapping operations: new map creation, relocalization in multiple maps, and map merging. We have devised how to interweave the elementary mapping stages to perform the multi-mapping operations robustly, accurately and efficiently. Among all the components of the system, it is relevant the map merging procedure that produces a seamless fusion of two maps with a common region. After the merge, the two merging maps are totally replaced by the new merged map. We propose the creation of a new map after tracking loss. It prevents the failure in exploratory trajectories in which relocalization cannot recover the camera tracking losses. 
\item A new criteria to declare the tracking lost in the case of poor camera pose observably. It is able to prevent erroneous pose graph optimizations in the loops that contain highly uncertain camera poses. 
\end{itemize}

We provide a quantitative experimental validation in the EuRoC datasets, in which ORBSLAM-Atlas achieves the best results to date for a global map after multiple sessions. In the monocular EuRoC difficult datasets, it greatly improves the coverage and localization error when compared with the single map ORBSLAM. Additionally, the system has proved outstanding robustness in dealing with dynamic scenes.

%-------------------------
\section{Related work}
In the literature, the multi-map capability has been researched as a component of collaborative mapping systems. The collaborating agents end up sending frames to a central server where the multiple mapping operations are performed. Foster et al. in \cite{forster2013collaborative} proposed for the first time this distributed architecture. In their approach, the agents send frames to the global server, however, they do not get information from the server to improve their local maps. The first system with bidirectional information flow, both from the agents to the server and from the server to the agents was C2TAM \cite{riazuelo2014c2tam} that is as an extension of PTAM \cite{klein2007parallel} to RGB-D sensors able to handle multiple maps in multiple robots. Morrison et al. in \cite{morrison2016moarslam} research a robust stateless client-server architecture for collaborative multiple-device SLAM. Their main focus is the software architecture, not reporting accuracy results. The recent work by Schmuck and Chli \cite{schmuck2017multi, schmuck2018ccm} proposes CCM-SLAM, a distributed multi-map for multiple drones, with bidirectional information flow, built on top of ORBSLAM. Our system is close to their central server because both are built on top of similar elementary mapping stages. They are focused on overcoming the challenge of a limited bandwidth and distributed processing in the monocular case, whereas our focus is building an accurate global map. According to their reported experiments in EuRoC Machine Hall datasets, our system is about 3 times more accurate in the monocular case. Additionally, our system displays robustness, processing accurately all the EuRoC datasets  both in stereo and monocular. 

The recent ORB-SLAMM \cite{daoud2018slamm} also proposes an extension of ORBSLAM2 to handle multiple maps in the monocular case. Their integration of the multiple maps is not so tight as ours, because their sub-maps are kept as separated entities, each having its own DBoW2 database. Additionally, their merge operation computes a link between the sub-maps but does not replace the merging sub-maps by the merged one.

%We also compare with VINS-Mono multi-session global map accuracy \cite{qin2018vins}, it is a monocular visual inertial odometry, in which loop correction is estimated by pose graph optimization. We report our ORBSLAM-Atlas multiple session stereo results on the same Machine Hall EuRoC datasets. ORBSLAM individual maps are 2 times more accurate than those of VINS-mono, because thanks to the map we are able to detect and BA process numerous high parallax observations. ORBSLAM-Atlas multi-session global map is able to retain the 2 times higher accuracy over the VINS-mono global map, because our map merging is able to detect and process the high parallax matches also in the multi-map and multi-session case.
%We also compare with VINS in its monocular inertial version, VINS-Mono \cite{qin2018vins}. It is a visual odometry system, in which loop correction is estimated by pose graph optimization. We compare VINS-Mono and ORBSLAM-Atlas accuracy in the multi-session processing of the Machine Hall EuRoC datasets. ORBSLAM-Atlas individual maps are 2 times more accurate than VINS-Mono. Because ORBSLAM-Atlas is able to detect and BA process numerous high parallax observations. ORBSLAM-Atlas multi-session global map retains the 2 times higher accuracy over the VINS-mono global map, because thanks to the map merging, is able to detect and process the high parallax matches also in the multi-map and multi-session case.

We also compare with VINS-Mono \cite{qin2018vins} in the multi-session processing of the Machine Hall EuRoC datasets. VINS-Mono is a visual odometry system, in which loop correction is estimated by pose graph optimization. As ORBSLAM-Atlas is able to detect and process with BA numerous high parallax observations, their individual maps are 2 times more accurate than those of VINS-Mono. ORBSLAM-Atlas multi-session global map retains the 2 times higher accuracy over the VINS-mono global map, because thanks to the map merging, it is able to detect and take profit from high parallax matches also in the multi-map and multi-session case.

The idea of adding robustness to track losses during exploration by means of map creation and fusion was firstly proposed by Eade and Drummond \cite{eadeBMVC2008} within a filtering approach. One of the first keyframe-based multi-map system was \cite{Castle2008}, where they proposed the idea of disconnected maps, however, the map initialization was manual, and the system was not able to merge or relate the different sub-maps. In the filtering EKF-SLAM approaches, where covariances are readily available, the camera was declared lost with a double criteria of threshold in the number of matches, and low camera localization error covariance \cite{williams2007real}. In the keyframe methods, the criterion was reduced just to the number of matches because the covariances are not computed. We propose to recover the double criteria, with a low cost proxy for the camera pose covariance, which comes from the Hessian of the camera-only pose optimization. This approximated covariance has been recently used in \cite{zhang2018perception} for active perception.

%------------------------------------------------------

\section{ORBSLAM-Atlas multi-map representation}
\begin{figure}
\centering
  \includegraphics[width=\columnwidth]{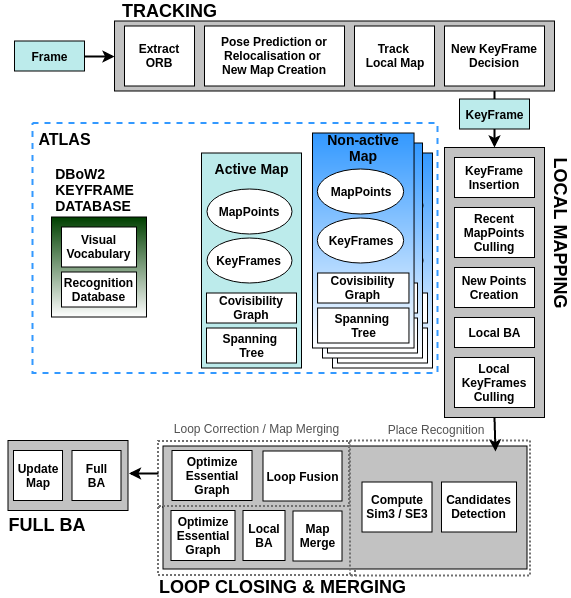}
  \caption{ORBSLAM-Atlas multi-map representation and workflow.}
  \label{fig:atlas_workflow}
\end{figure}

We call the new multiple map representation \emph{atlas}, from now on, we will use the name \emph{map} to designate each of the atlas sub-maps. Next subsections detail the atlas structure and the criteria to determine when a new map has to be created.

\subsection{Multi-map representation}
The atlas (Fig.\,\ref{fig:atlas_workflow}) is composed of a virtually unlimited number of maps, each map having its own keyframes, map points, covisibility graph and spanning tree. Each map reference frame is fixed in its first camera, and it is independent of the other maps references as in ORBSLAM. The incoming video updates only one map in the atlas, we call it the \emph{active map}, we refer to the rest of the maps as \emph{non-active maps}. The atlas also contains a \emph{unique for all the maps} DBoW2 recognition database that stores all the information to recognize any keyframe in any of the maps. 

Our system has a single place recognition stage to detect common map regions, if both of them are in the active-map, they correspond to a loop closure, whereas if they are in different maps, they correspond to a map merge. 

%The loop correction makes a pose graph optimization that provides the initial guess from where the next local BA efficiently converges. 

%The local BA produces a local map that seamless fuses the two loop ends, finally a full BA further refines the total map. In contrast, the map merging the stages are scheduled differently. There is an initial alignment of the merged sub-map into the active map, then a local BA computes the seamless fusion in the merging area, and finally a pose graph optimization propagates the correction to the resulting merged map that replaces the active map.

\subsection{New map creation criteria}
When the camera tracking is considered lost, we try to relocalize in the atlas. If the relocalization is unsuccessful for a few frames, the active map becomes a non-active map and is stored in the atlas. Afterwards, a new map initialization is launched according to the algorithms described in \cite{mur2017orb} and \cite{mur2015orb}. 
 
To determine if the camera is on track, we heuristically propose two criteria that have to be fulfilled, otherwise, the camera is considered lost:
\paragraph{Number of matched features}: the number of matches between the current frame and the points in the local map is above a defined threshold. 
\paragraph{Camera pose observability}: if the geometrical conditioning of the detected points is poor, then camera pose will not be observable and the camera localization estimate will be inaccurate.

\begin{figure}[ht]
  \centering
  \begin{subfigure}{\columnwidth}
    \centering
    \includegraphics[width=0.7\columnwidth,trim={0 0 0 1cm},clip]{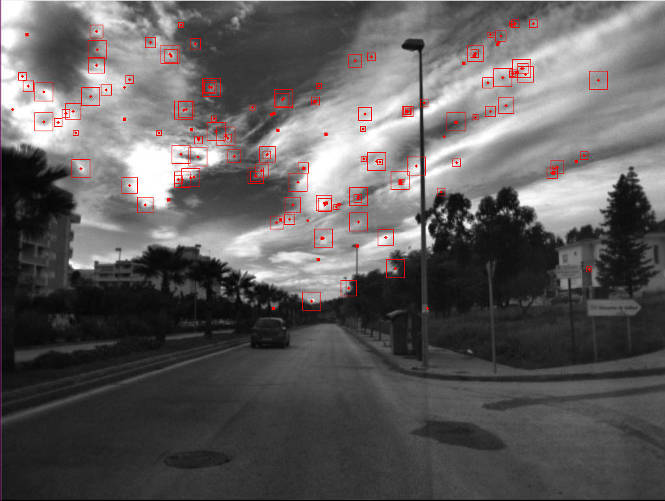}
    \caption{}
  \end{subfigure}
  \begin{subfigure}{\columnwidth}
    \centering
    \includegraphics[width=\columnwidth]{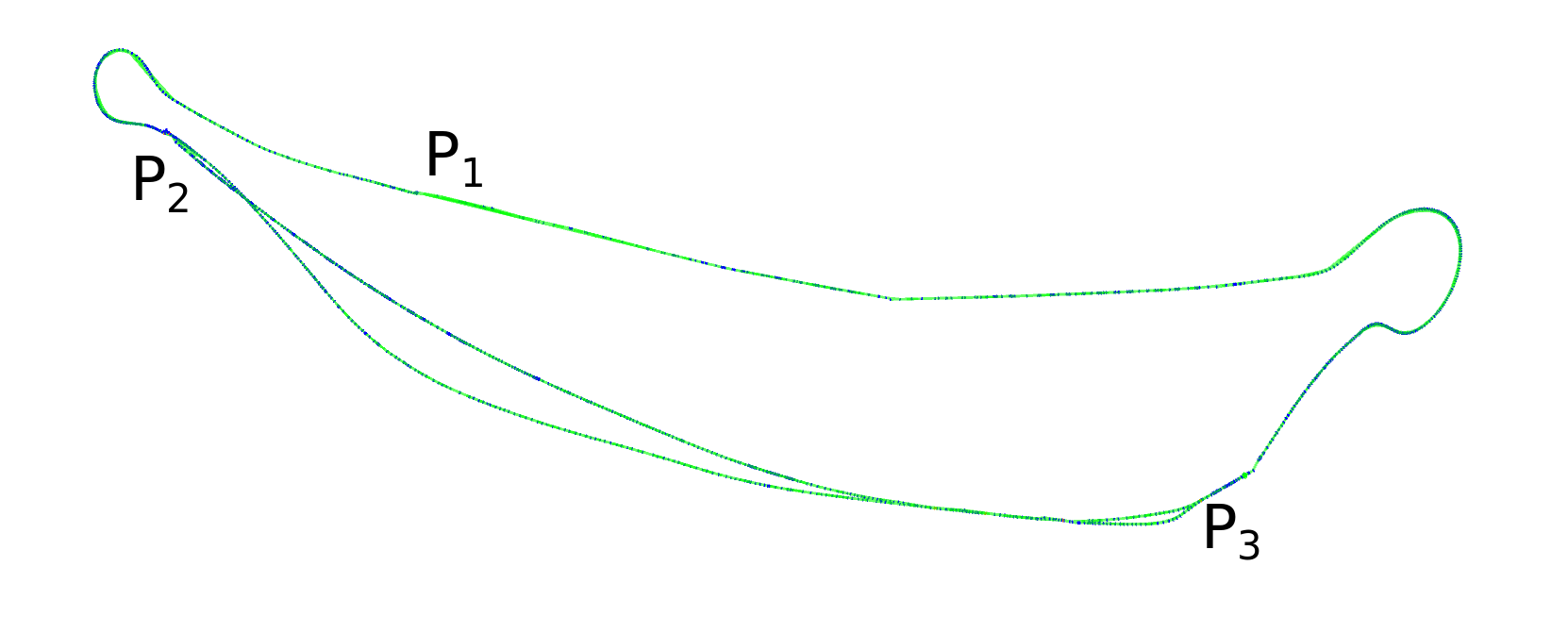}
    \caption{}
  \end{subfigure}
 \begin{subfigure}{\columnwidth}
    \centering
    \includegraphics[width=\columnwidth]{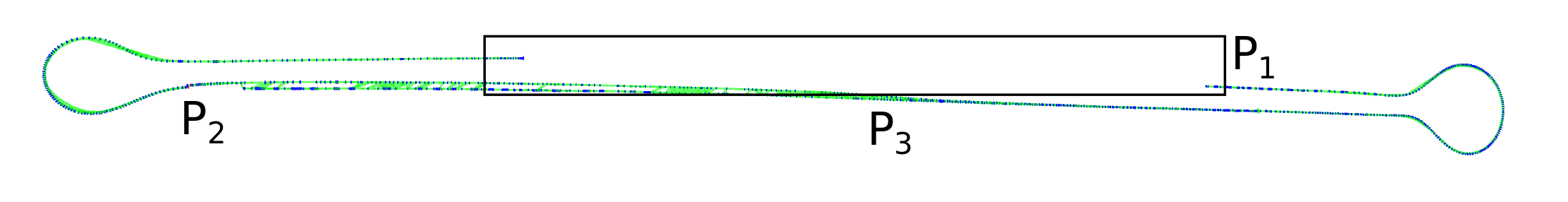} 
    \vspace*{-0.6cm}
    \caption{}
  \end{subfigure}
  \caption{Example of mapping accuracy improvement due to observability criterion. (a) Frame where most of the matched points are far from the camera. The number of points criterion is fulfilled but not the observability criterion, and the camera translation is inaccurately estimated. The image corresponds to the region marked as $P_1$ in the maps below. (b) Camera trajectory without observability criterion. Two loop closures were detected at $P_2$ and $P_3$, due to the inaccurate camera poses around P1, the pose graph optimization fails to produce an accurate correction. (c) Camera trajectory with observability criterion. The camera poses in the rectangle in the $P_1$ region area are excluded. When the low observability region is left, a second map is created. When $P_2$ is reached, the place recognition fires, and the two maps are merged into a single map. At $P_3$ a loop closing is detected applying the corresponding correction. The final global map has fewer localized frames but they are more accurate.}
  \label{fig:lost_observability}
\end{figure}

Figure\,\ref{fig:lost_observability} displays an example from the Malaga datasets \cite{blanco2014malaga}, where the usage of the covisibility criteria, combined with the multiple mapping produces a dramatic improvement in the mapping accuracy. A number of points over the threshold are matched in the image, however, they correspond to distant map points, hence the camera translation is estimated inaccurately. Without the observability criterion, the loop closure correction computed by the pose graph optimization is inaccurate due to the poor accuracy of the relative  translations included in the loop. Whereas if the observability criterion is used, those uncertain keyframes are removed from the map, the map is fragmented but ORBSLAM-Atlas is able to merge all the sub-maps in an accurate global map.

\subsection{Camera pose observability}

We estimate the observability from the camera pose error covariance. We assume the map points are perfectly estimated because the real-time operation cannot afford to compute the covariance for the map points per each frame. The measurement information matrix, $\mathbf{\Omega}_{i,j}$, coding the uncertainty for the observation, $\mathbf{x}_{i,j}$, of the map point $j$ in camera $i$. It is tuned proportional to image resolution scale where the image FAST point has been detected. The uncertainty of the camera $i$ is estimated with the $m_i$ points, where $m_i$ is the number of points in the camera $i$ matched with the map points.

We estimate the 6 d.o.f camera pose as the $\mathbf{\hat{T}}_{i,w} \in \mbox{SE}(3)$ transformation. Additionally, we code its uncertainty by means of the unbiased Gaussian vector of 6 parameters $\boldsymbol{\varepsilon}_{i}$ that defines the Lie algebra approximating $\mathbf{T}_{i,w}$ around $\mathbf{\hat{T}}_{i,w}$:
\begin{eqnarray*}
  \mathbf{T}_{i,w} &=& \text{Exp}\left({\boldsymbol{\varepsilon}_{i}}\right)\oplus \mathbf{\hat{T}}_{i,w}\\
  \boldsymbol{\varepsilon}_{i} &=& \begin{pmatrix}
  x & y & z & \omega_{x} & \omega_{y} & \omega_{z}
  \end{pmatrix} \sim \mathcal{N}(0,\mathbf{C}_{i}) \\
  \mathbf{H}_{i} &\simeq& \sum_{j=1}^{m_i} \mathbf{J}^{\intercal}_{i,j} \mathbf{\Omega}_{i,j} \mathbf{J}_{i,j}\\
  \mathbf{C}_{i} &=& \mathbf{H}_{i}^{-1}
\end{eqnarray*}
where $\text{Exp}: \mathbb{R}^6 \rightarrow \text{SE}(3)$  directly maps from the parameters space $\boldsymbol{\varepsilon}_i \in \mathbb{R}^6$ to the Lie group $\text{SE}(3)$.
The covariance matrix $\mathbf{C}_i$ codes the camera estimation accuracy and $\mathbf{J}_{i,j}$ is the Jacobian matrix for the camera pose measurement due to the observation of the map point $j$ in the camera $i$. As translation is the weakly observable magnitude,  we propose to use in the criterion only the $\mathbf{C}_i$ diagonal values corresponding to the translation error:
\begin{eqnarray}
  \max\left(\sigma_x, \sigma_y, \sigma_z\right) &<&\sigma_{th}^t \\\label{eq:translation_error_test}
  \left[\begin{array}{cccccc}\sigma_x^2 & \sigma_y^2&\sigma_z^2 &\sigma_{\omega_x}^2 & \sigma_{\omega_y}^2&\sigma_{\omega_z}^2 \end{array}\right] &=&\mbox{diag}\left(\mathbf{C}_{i}\right) \nonumber
\end{eqnarray}
%\sqrt{\sigma_{\omega_x}^2+\sigma_{\omega_y}^2+ \sigma_{\omega_z}^2} &<&\sigma_{th}^o \\
%\end{eqnarray}

\subsection{Relocalization in multiple maps}
\label{sec_relocalization}
If camera tracking is lost, we use the frame to query the atlas DBoW database. This single query is able to find the more similar keyframe in any of the maps. Once we have the candidate keyframe, map, and the putative matched map points, we perform the relocazation following \cite{mur2015orb}. It includes robustly estimating the camera pose by a first PnP and RANSAC stage, followed by a guided search for matches and a final non-linear camera pose-only optimization.

%-----------------------------------------------

\section{Seamless map merging}
\label{sec_map_merging}
For detecting map merges we use the ORBSLAM place recognition stage. It enforces repeated place recognition for three keyframes connected by the covisibility graph in order to reduce the false positive risk. Additionally, in the merging process, the active map swallows the other map where the common regions have been found. Once the merging is complete the merged map completely replaces the two merging maps. When necessary, we will use the $a$, $s$, and $m$ subindexes to refer to the active, swallowed and merged maps respectively.

%Initially the recognition 

\begin{enumerate}
    \item \textbf{Detection of common area between two maps}. The place recognition provides two matching keyframes, $K_a$ and $K_s$ and a set of putative matches between points in the two maps $M_a$ and $M_s$
    \item \textbf{Estimation of the aligning transformation}. It is the transformation, $\mbox{SE}(3)$ in stereo or $\mbox{Sim}(3)$ in monocular, that aligns the world references of the two merging maps. We compute an initial estimation combining Horn method \cite{horn1987closed} with RANSAC, from the putative matches between $M_a$ and $M_s$ map points. We apply the estimated transformation to $K_s$ for a guided matching stage, where we match points of $M_a$ in $K_s$, from which we eventually estimate $\mathbf{T}_{W_a, W_s}$ by non-linear optimization of the reprojection error.
    \item \textbf{Combining the merging maps}. We apply $\mathbf{T}_{W_a, W_s}$ to all the keyframes and map points in $M_s$. Then, we detect duplicated map points and fuse them, what yields map points observed both from keyframes in $M_s$ and $M_a$. Afterwards, we combine all $M_s$ and $M_a$ keyframes and map points into  $M_m$. Additionally, we merge the  $M_s$ and $M_a$ spanning trees and  covisibility graphs into the spanning tree and  covisibility graph of  $M_m$.
    \item \textbf{Local BA in the welding area}. It includes all the keyframes covisible with $K_a$ according to $M_m$ covisibility graph. To fix the gauge freedoms the keyframes that were fixed in $M_a$ are kept fixed in the local BA, whereas the rest of the keyframes are set free to move during the non-linear optimization. We apply a second duplicated point detection and fusion stage updating the $M_m$ covisibility graph.
    \item \textbf{Pose graph optimization}. Finally, we launch a pose graph optimization of $M_m$.
\end{enumerate}

The merging runs in a thread in parallel with the tracking thread, the local mapping thread, and occasionally a global bundle adjustment thread (Fig.\ref{fig:atlas_workflow}.) Before starting the merging, the local mapping thread is stopped to avoid the addition of new keyframes in the atlas. If a global bundle adjustment thread is running, it is also stopped because the spanning tree on which the BA is operating is going to be changed. The tracking thread is kept running on the old active map to keep the real-time operation. Once the map merging is finished, we resume the local mapping thread. The global bundle adjustment, if it has been stopped, is relaunched to process the new data.

%-------------------------------------

\section{Experiments}
\label{sec_expeiments}

The quantitative evaluation has been made in the EuRoC datasets \cite{burri2016euroc}. To score the results we compute the RMS ATE (Absolute Translation Error) in meters for all the frames in the sequences as proposed in \cite{sturm2012benchmark}. To factor out the non-deterministic nature of the multi-threading execution, we run each experiment 5 times and report the average or median values. 
The qualitative evaluation was done in monocular for a hand-held camera traversing a densely populated corridor where occlusions and tracking losses are frequent. For a general overview of the experiments see the accompanying video.

\subsection{Multiple map performance}

\begin{table*}[ht]
\resizebox{\textwidth}{!} {
\centering
\begin{tabular}{|l|c|c|c|c|c|c||c|c|c|c|c|c|}
\hline
\multirow{2}{*}{} & \multicolumn{3}{c|}{\begin{tabular}{c}ORBSLAM-Atlas \\ Monocular \end{tabular}} & \multicolumn{3}{c||}{\begin{tabular}{c} ORBSLAM \\ Monocular \end{tabular}} & \multicolumn{3}{c|}{\begin{tabular}{c} ORBSLAM-Atlas \\ Stereo \end{tabular}} & \multicolumn{3}{c|}{\begin{tabular}{c} ORBSLAM2 \\ Stereo \end{tabular}} \\ \cline{2-13}
                & ATE (m)& Cover (\%) & \# Maps & ATE (m) & Cover (\%) & \# Maps & ATE (m) & Cover (\%)& \# Maps & ATE (m) & Cover (\%) & \# Maps \\ \hline
V1\_03 & \textbf{0.106} & \textbf{90.74} & 2 & 0.132 & 10.32 & 1 & 0.051 & 100 & 1 & \textbf{0.046} & 100 & 1 \\ \hline
V2\_03 & \textbf{0.093} & \textbf{70.74} & 2 & 0.146 & 15.71 & 1 & \textbf{0.218} & \textbf{94.55} & 5 & 0.316 & 89.21 & 1 \\ \hline
\end{tabular}}
\caption{Performance on the difficult Vicon Room EuRoC datasets. RMS ATE in meters. Median values after 5 runs.}
\label{table:euroc_challenged_comp}
\end{table*}

\begin{figure}
    \centering
    \begin{subfigure}{\columnwidth}
        \includegraphics[width=\columnwidth]{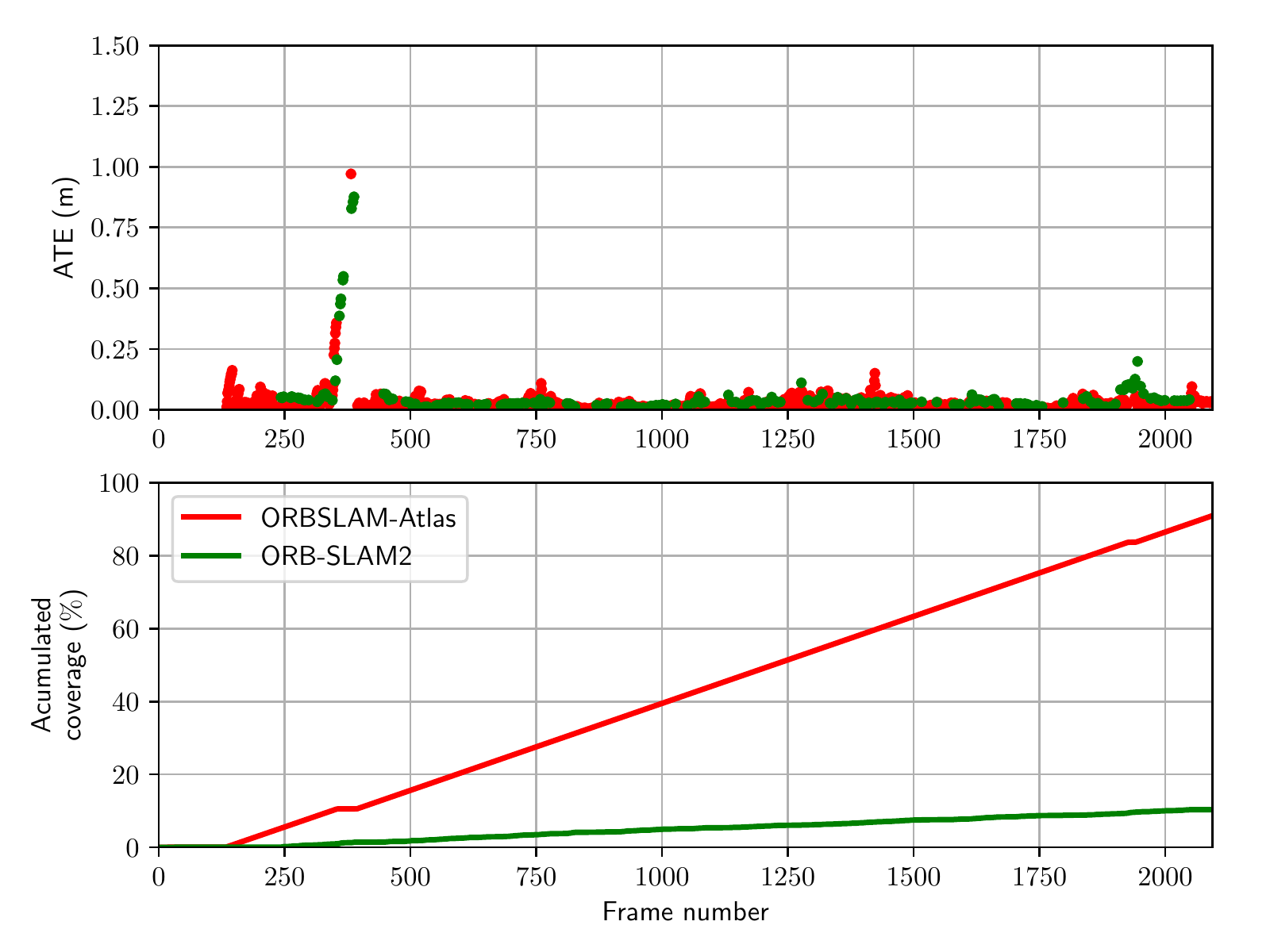}
        \caption{V1\_03 in monocular}
    \end{subfigure}
    \begin{subfigure}{\columnwidth}
        \includegraphics[width=\columnwidth]{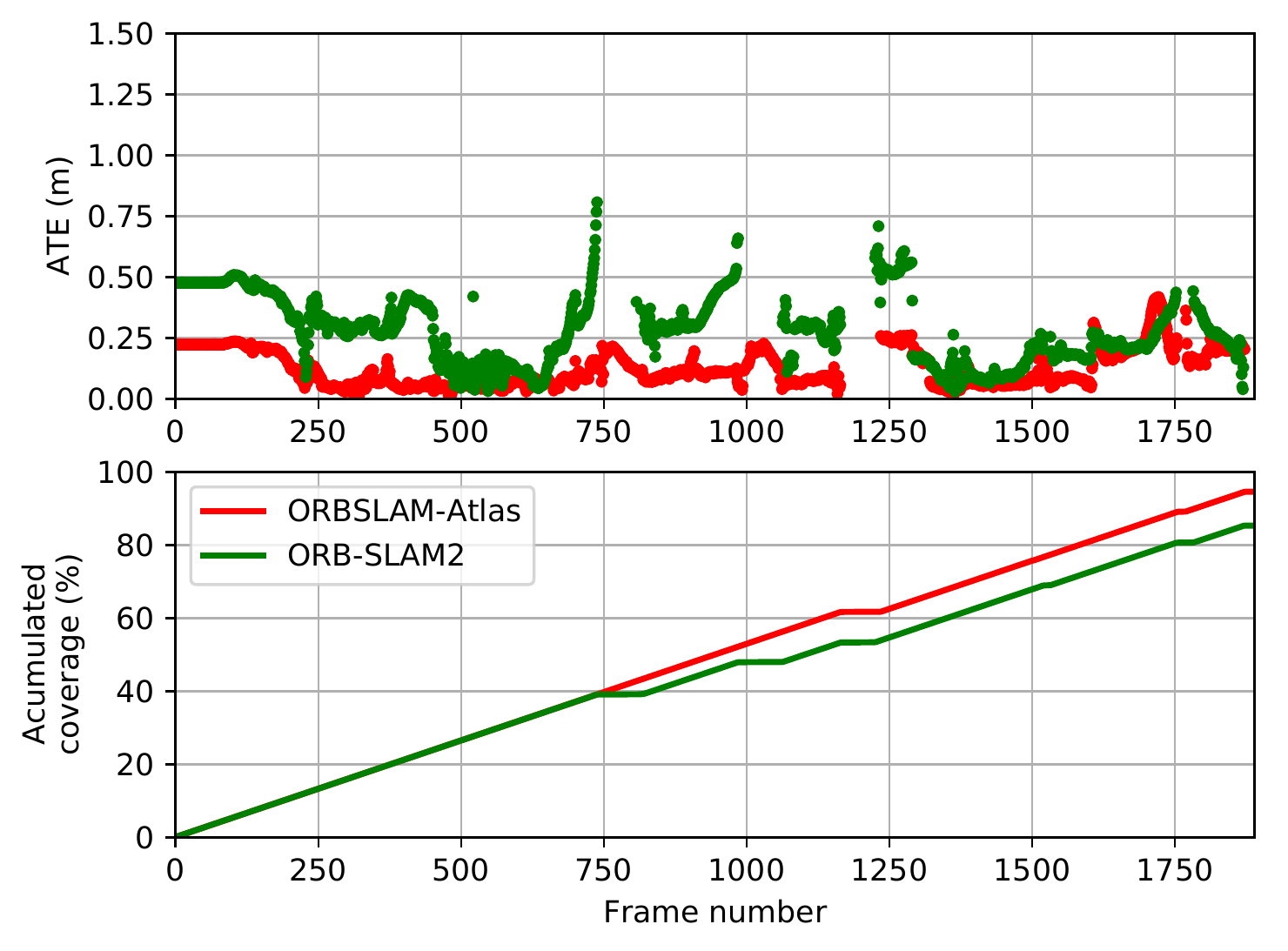}
        \caption{V2\_03 in stereo}
    \end{subfigure}
    \caption{ATE (m) per each localized frame in the sequence, and accumulated coverage (\%). Out of the 5 runs, it is represented the one that gets the median RMS ATE. Best viewed in color.}
    \label{fig:euroc_challenged_sequences}
\end{figure}

We focus our quantitative evaluation on the EuRoC V1\_03\_difficult and V2\_03\_difficult datasets because ORBSLAM2 stereo \cite{mur2017orb} or ORBSLAM monocular \cite{mur2017visual} reported them as failure due to a coverage below 90\,\%. Coverage is defined as the fraction of localized frames with respect to the total number of ground truth frames in the dataset. The differences in performance in the rest of the datasets are negligible because ORBSLAM-Atlas never lost track, and hence never used more than a single map. 

Table\,\ref{table:euroc_challenged_comp} reports the quantitative comparison, see also Figure\,\ref{fig:euroc_challenged_sequences}. We have made new experiments with ORBSLAM to report both the RMS ATE and the coverage. Thanks to the multi-maps, ORBSLAM-Atlas is able to significantly boost the coverage from 10-15\,\% to  70-90\,\%, with an RMS ATE lower than ORBSLAM.  

In the stereo case, in V1\_3 the differences between ORBSLAM2 and ORBSLAM-Atlas are negligible. In contrast, in V2\_3 ORBSLAM-Atlas produces 5 intermediate maps that eventually are merged in a global map able to achieve around 95\,\% coverage and an RMS ATE lower that ORBSLAM2.

\subsection{Multi-session performance}
\begin{table}
\centering
\resizebox{\columnwidth}{!} {
 \begin{tabular}{|l|c|c|c|}
 \hline
            & \begin{tabular}{c}ORBSLAM-Atlas\\stereo\end{tabular} & \begin{tabular}{c}VINS\\stereo\end{tabular} & \begin{tabular}{c}VINS\\Mono Inertial\end{tabular} \\ \hline
    V1\_01  & \textbf{0.036} & 0.550 &  0.068    \\ \hline  
    V1\_02  & \textbf{0.022} & 0.230 &  0.084    \\ \hline  
    V1\_03  & \textbf{0.051} &   X   &  0.190    \\ \hline  
    V2\_01  & \textbf{0.034} & 0.230 &  0.081    \\ \hline  
    V2\_02  & \textbf{0.028} & 0.200 &  0.150    \\ \hline  
    V2\_03  & \textbf{0.218} &   X   &  0.220    \\ \hline 
    MH\_01  & \textbf{0.036} & 0.540 &  0.120    \\ \hline
    MH\_02  & \textbf{0.021} & 0.460 &  0.120    \\ \hline 
    MH\_03  & \textbf{0.026} & 0.330 &  0.130    \\ \hline
    MH\_04  & \textbf{0.103} & 0.780 &  0.180    \\ \hline
    MH\_05  & \textbf{0.054} & 0.500 &  0.210    \\ \hline \hline
    \begin{tabular}{r}muliple-session\\MH\_01-MH\_05\end{tabular}\hspace*{-0.3cm} & \multicolumn{1}{|r|} {\textbf{0.086}} &  -  & \multicolumn{1}{|r|} {0.210}  \\ \hline
\end{tabular}}
\caption{Multiple-session performance on EuRoC datasets. We report the results of the individual mapping sessions, and the global multi-session map after the sequential processing of datasets MH\_01 to MH\_05. Reported RMS ATE (m) are median values after 5 runs.}
\label{table:euroc_muti_session_vs_mono_session}
\end{table}

Table \ref{table:euroc_muti_session_vs_mono_session} displays the RMSE ATE for all the datasets in EuRoC, which are processed individually. We also report the global multi-session map after processing the five Machine Hall datasets (MH\_01 to MH\_05) sequentially for ORBSLAM-Atlas and VINS-Mono. For VINS-Mono and VINS-Stereo we verbatim quote the values reported by the authors in \cite{qin2018vins,qin2019general}. Trajectories have been aligned by means of $\mbox{SE}(3)$ transformations. 

We can conclude that our individual session maps are more accurate than those of VINS-Mono or VINS-Stereo. We conjecture that ORBSLAM-Atlas can detect numerous high parallax observations and process them with non-linear BA, and hence is more accurate.
The same accuracy advantage between ORBSLAM-Atlas and VINS-Mono is retained in the multiple session case, what proves that ORBSLAM-Atlas is able to detect and exploit the high parallax matches also among the multiple maps, and in the multiple session operation. 

In table \ref{table:comparation_multiagent_prec}, we compare  with respect to CCM-SLAM\cite{schmuck2017multi, schmuck2018ccm}, which is a centralised collaborative monocular SLAM system where the agents compute a local map and  send frames to the central server in order to build a global map. In the experiment reported in their paper, CCM-SLAM is launched with three agents, each of them processes, in parallel, a sequence of the  EuRoC Machine Hall experiment (MH\_01, MH\_02 and MH\_03), and the server processes all the information from the three sequences in the global map. The reported RMS ATE is computed with respect to the ground truth after a $\mbox{Sim}(3)$ alignment. We verbatim quote the values as reported by the authors in \cite{schmuck2018ccm}. We have processed the MH\_01, MH\_02 and MH\_03 datasets sequentially in a multi-session manner with ORBSLAM-Atlas to obtain a global map. We have made the monocular mapping with the corresponding $\mbox{Sim}(3)$ alignment. We have also made the stereo mapping, hence we can recover the scale, and report the RMS ATE after $\mbox{SE}(3)$ alignment. We can conclude that our global map is more accurate than CCM-SLAM in the monocular case. Additionally, the stereo case also shows better accuracy with the advantage that we estimate the scene real scale.

\begin{table}
\centering
\resizebox{\columnwidth}{!} {
\begin{tabular}{|l|c|}
\hline
                       & Global map RMS ATE (m) \\ \hline
CCM-SLAM (Mono*)       &         0.077          \\ \hline
ORBSLAM-Atlas (Mono*)  & \textbf{0.024}         \\ \hline\hline
ORBSLAM-Atlas (Stereo) &         0.035          \\ \hline
\end{tabular}}
\caption{RMS ATE (m) in the EuRoC Machine Hall (MH\_01, MH\_02 and MH\_03). * indicates that the aligning transformation prior to ATE computation includes a scale correction. The reported values are the average after 5 runs to make them comparable with results reported in \cite{schmuck2018ccm}.}
\label{table:comparation_multiagent_prec}
\end{table}

\begin{figure}
    \centering
    % \begin{subfigure}{\columnwidth}
        \includegraphics[width=\columnwidth]{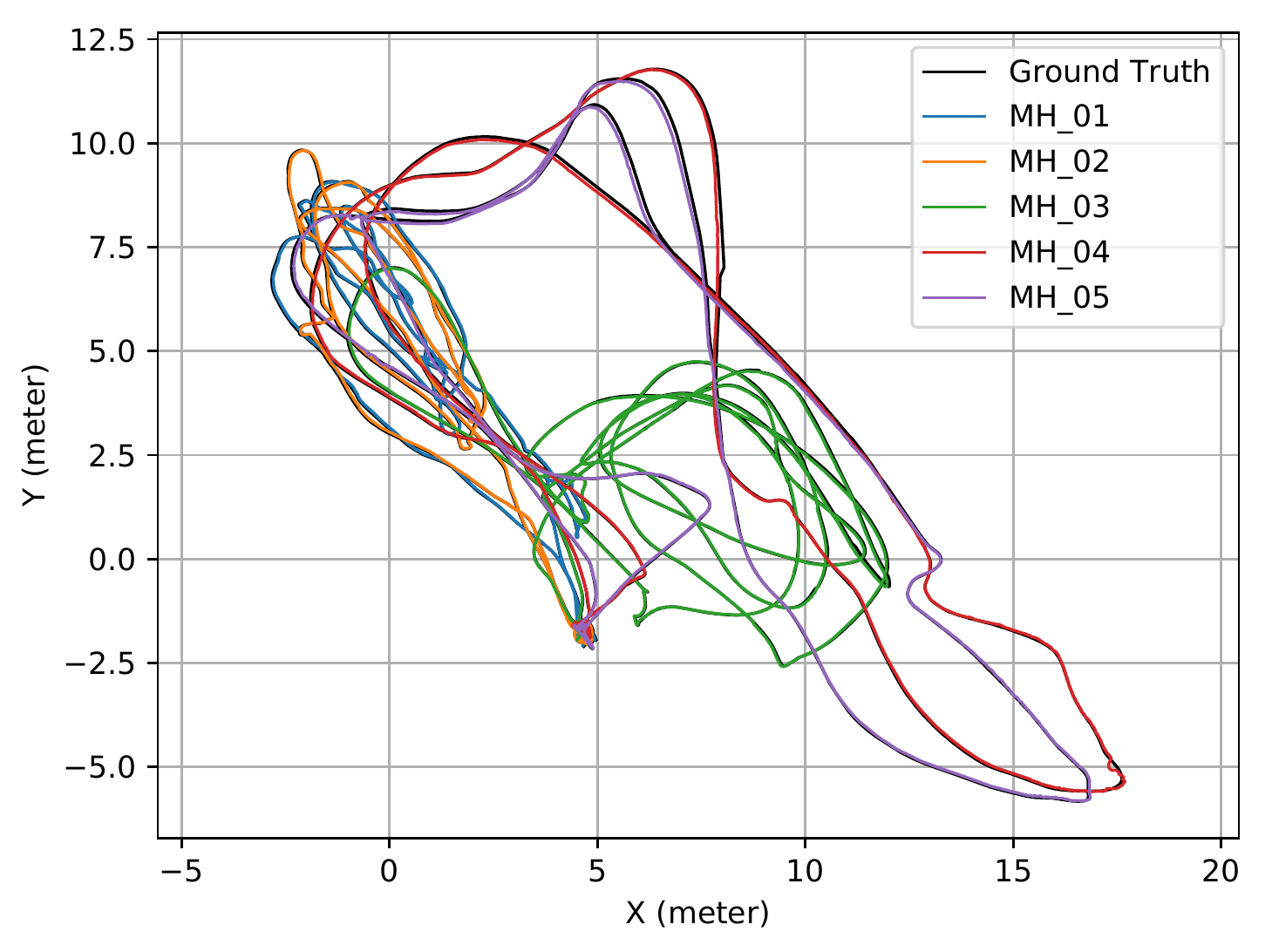}
    % \end{subfigure}
    % \begin{subfigure}{\columnwidth}
    %     \includegraphics[width=\columnwidth]{images/EuRoC_TrajectoriesMH.png}
    %     \caption{}
    % \end{subfigure}
    \caption{Trajectories after processing  Machine Hall datasets MH\_01-MH\_05 sequentially as multiple sessions with ORBSLAM-Atlas stereo (top view). Aligned with ground truth by means of global $\mbox{SE}(3)$ transformation. Best viewed in color.}
    \label{fig:euroc_mh_trajectories}
\end{figure}

\subsection{Mapping in dynamic scenes}

\begin{figure}
        \includegraphics[width=\columnwidth]{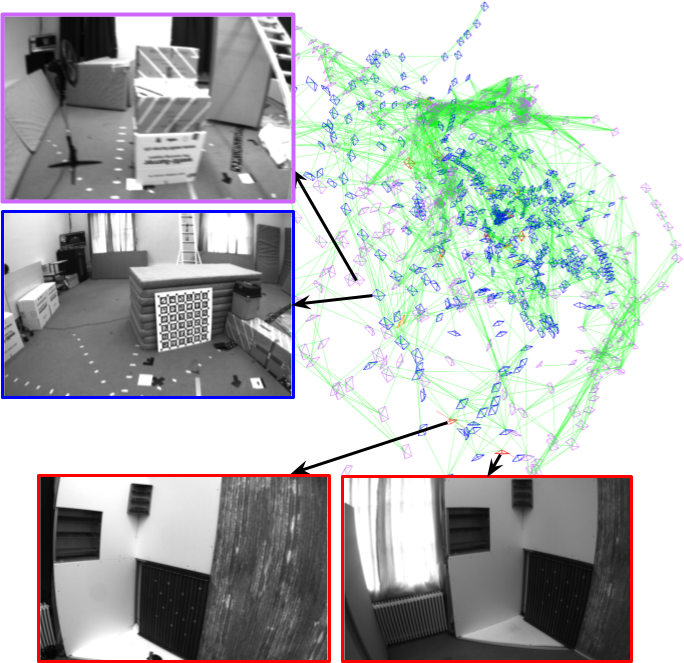}
    \caption{Keyframes of the Vicon room global map. The map contains the two experiences corresponding to the two versions of the room. All the keyframes of the global map are displayed, the purple keyframes correspond to V1\_XX, the blue ones to V2\_XX. Two keyframes close in space but corresponding to different experiences are displayed at top left corner. The two bottom keyframes corresponds to the merging keyframes.}
    \label{fig:two_experiences}
\end{figure}

In the accompanying video, we provide a qualitative evaluation in a fast dynamic scene, in which a monocular hand-held camera images a densely populated environment. ORBSLAM-Atlas is able to produce a global map for the whole plant corridor. Several intermediate maps have been spawned to survive to camera tracking losses. 

To provide quantitative evaluation, we have processed the whole EuRoC dataset in a multi-session manner, feeding the 11 stereo videos in sequence: MH\_01, MH\_02, MH\_03, MH\_04, MH\_05, V1\_01, V1\_02, V1\_03, V2\_01, V2\_02, V2\_03, without providing any additional information to the system. After the 11 sessions, the system has been able to identify three different maps. The first map corresponds to the five sequences of the Machine Hall. The second map corresponds to V1\_01, V1\_02, V1\_03, V2\_01 and V2\_02. Experiments V1\_XX and V2\_XX were grabbed in the same room,  however experiments V2\_XX were made 112 days later than V1\_XX, the distribution of the furniture was changed, and the ground truth reference was moved as well. Our system is able to merge the maps corresponding to the two versions of the room because of the common elements, which mainly correspond to the floor and the elements fixed to the walls, such as the door, the windows or the radiators. The third map corresponds to sequence V2\_03 that, due to the fast camera motion, our system is unable to merge with the second map.

The merged map of the Vicon room is interesting because it displays the lifelong capabilities of our system. The same map is able to jointly consider the two different experiences of the same room. There are some pairs of keyframes localized close to each other in the map, however they image two different version of the room (see Fig.\,\ref{fig:two_experiences}), and are not connected in the covisibility graph. Thanks to the accuracy of the place recognition and the feature matching, the system never gets confused with the different versions of the room, but reuses the keyframes when the camera observes common scene areas. In Table \ref{table:euroc_comp_maps_size}  we report the global map error, and the map size in terms of the number of keyframes and the number of map points. In the case of the Machine Hall, there is a reduction in the number of keyframes (82\,\%) and keypoints (52\,\%) of the global map with respect to the individual maps. The reduction is proportional to the common areas between the maps (see Fig.\,\ref{fig:euroc_mh_trajectories}). In the case of the Vicon room, this reduction is only slightly smaller (89\,\% for KF and 60\,\% for KP) despite the drone trajectories are close to each other. There is no bigger reduction because the global map has to represent the two versions of the room. The global reference for the ground truth in the two rooms was different, for this reasons, to compute the RMS ATE we have made two $\mbox{SE}(3)$ alignments, one for the V1 room frames and other to the V2 frames.

\begin{table}
\centering
\resizebox{\columnwidth}{!} {
\begin{tabular}{|l|c|c|c|}
\hline
Dataset & \# KF & \# MP  & RMSE ATE (m) \\\hline
MH\_01  &  481  & 10,199 &     0.035    \\ \hline
MH\_02  &  430  & 16,504 &     0.018    \\ \hline
MH\_03  &  442  & 19,947 &     0.028    \\ \hline
MH\_04  &  316  & 18,943 &     0.119    \\ \hline 
MH\_05  &  373  & 21,203 &     0.060    \\ \hline 
\multicolumn{1}{|r|}{Total Size} & \begin{tabular}{c}2,042\\(100\,\%) \end{tabular}  &\begin{tabular}{c}86,796\\(100\,\%) \end{tabular} & -\\ \hline
\begin{tabular}{l}MH\_01+MH\_02+MH\_03+\\MH\_04+MH\_05 \end{tabular} & \begin{tabular}{c} 1,666\\(82\,\%) \end{tabular} & \begin{tabular}{c} 45,660\\(53\,\%) \end{tabular} & 0.086  \\ \hline
V1\_01  &  112  &  7,610 &     0.035    \\ \hline
V1\_02  &  145  &  9,682 &     0.020    \\ \hline
V1\_03  &  228  & 13,291 &     0.048    \\ \hline
V2\_01  &  109  &  7,902 &     0.037    \\ \hline 
V2\_02  &  292  & 16,081 &     0.035    \\ \hline 
\multicolumn{1}{|r|}{Total Size} & \begin{tabular}{c} 886\\(100\,\%) \end{tabular} & \begin{tabular}{c} 54,566\\(100\,\%) \end{tabular}  & -\\ \hline 
\begin{tabular}{l} V1\_01+V1\_02+V1\_03+\\
V2\_01+V2\_02\end{tabular} & \begin{tabular}{c} 791\\(89\,\%) \end{tabular} & \begin{tabular}{c} 32,920\\(60\,\%) \end{tabular} & 0.040\\ \hline
V2\_03  &  270  & 13,683 &     0.218    \\ \hline
\end{tabular}}
\caption{Multiple-map in a dynamic scene. ORBSLAM-Atlas stereo identifies 3 different maps.
Comparison of the individual session mapping with respect to the multi-session mapping. Median values after 5 runs.}
\label{table:euroc_comp_maps_size}
\end{table}

% -------------------------------------------------

\subsection{Computing Time}
We have evaluated our ORBSLAM-Atlas algorithm in an Intel Core i7-7700 (four cores @ 3.6 GHz) desktop computer with 32GB RAM. We focus on the V2\_03 EuRoC dataset in stereo, the frame rate is 20\,Hz. We can achieve real time in the tracking thread with an average processing time of $\approx42\,\mbox{ms}$. The local mapping, running in a parallel thread, typically consumes $\approx78\,\mbox{ms}$ per keyframe. Place recognition takes $\approx10\,\mbox{ms}$ to compute the aligning transformation and map merging takes $\approx670\,\mbox{ms}$. In any case, as map merging runs in a parallel thread, it does not interfere the real-time tracking thread. Tracking operates on the unmerged map until merging is finished, and then the unmerged map is substituted by the merged one.

\section{Conclusions}
\label{sec_concluision}

% -------------------------------------------------
We have presented ORBSLAM-Atlas a multi-map system able to bring the outstanding qualities of the single map ORBSLAM to the multiple map arena. It is able, not only to robustly detect wide-baseline matches between the sub-maps but also, to include them in the subsequent non-linear optimizations to yield accurate estimations for the cameras and the map. The resulting multi-map system is more robust because it is able to survive to the tracking losses in exploratory trajectories, and more general because it naturally can handle multi-session operation.

The experimental validation in the EuRoC datasets has revealed that ORBSLAM-Atlas can report the best results to date for a global map after multi-sessions, and for the coverage and error in the EuRoC difficult datasets sing purely monocular vision. Additionally, the system has proved outstanding robustness in dealing with dynamic scenes.

\bibliographystyle{IEEEtran}
\bibliography{./references,IEEEabrv}

\end{document}